\documentclass[iicol,pdflatex,sn-mathphys-num]{sn-jnl}

\usepackage{graphicx}%
\usepackage{multirow}%
\usepackage{amsmath,amssymb,amsfonts}%
\usepackage{amsthm}%
\usepackage{mathrsfs}%
\usepackage[title]{appendix}%
\usepackage{xcolor}%
\usepackage{textcomp}%
\usepackage{manyfoot}%
\usepackage{booktabs}%
\usepackage{algorithm}%
\usepackage{algorithmicx}%
\usepackage{algpseudocode}%
\usepackage{listings}%
\usepackage{longtable}
\usepackage{array}
\usepackage{geometry}
\usepackage{caption}
\usepackage{algorithm}
\usepackage{algpseudocode}

\theoremstyle{thmstyleone}%
\theoremstyle{thmstyletwo}%

\theoremstyle{thmstylethree}%

\begin{document}

\title[Article Title]{Enhancing Cocoa Pod Disease Classification via Transfer Learning and Ensemble Methods: Toward Robust Predictive Modeling}

\author{\fnm{Devina} \sur{Anduyan}}
\author{\fnm{Nyza} \sur{Cabillo}}
\author{\fnm{Navy} \sur{Gultiano}}
\author{\fnm{Mark Phil} \sur{Pacot}}

\affil{\orgdiv{Department of Computer Science, College of Computing and Information Sciences}, \orgname{Caraga State University}, \orgaddress{\city{Butuan City}, \postcode{8600}, \country{Philippines}}}

\abstract{This study presents an ensemble-based approach for cocoa pod disease classification by integrating transfer learning with three ensemble learning strategies: Bagging, Boosting, and Stacking. Pre-trained convolutional neural networks, including VGG16, VGG19, ResNet50, ResNet101, InceptionV3, and Xception, were fine-tuned and employed as base learners to detect three disease categories: Black Pod Rot, Pod Borer, and Healthy. A balanced dataset of 6,000 cocoa pod images was curated and augmented to ensure robustness against variations in lighting, orientation, and disease severity. The performance of each ensemble method was evaluated using accuracy, precision, recall, and F1-score. Experimental results show that Bagging consistently achieved superior classification performance with a test accuracy of 100\%, outperforming Boosting (97\%) and Stacking (92\%). The findings confirm that combining transfer learning with ensemble techniques improves model generalization and reliability, making it a promising direction for precision agriculture and automated crop disease management.
}


\maketitle

\section{Introduction}\label{sec1}
Cocoa (\textit{Theobroma cacao}), a high-value tropical crop essential to global chocolate production, is severely threatened by diseases such as Black Pod Rot and Pod Borer. In the Philippines, it is recognized under Republic Act 7900 as a priority crop, yet disease outbreaks continue to impact yield and quality. Accurate and timely detection of these diseases is critical for effective management.

\begin{figure*}[t]
	\centering
	\includegraphics[width=\textwidth,height=8cm,keepaspectratio]{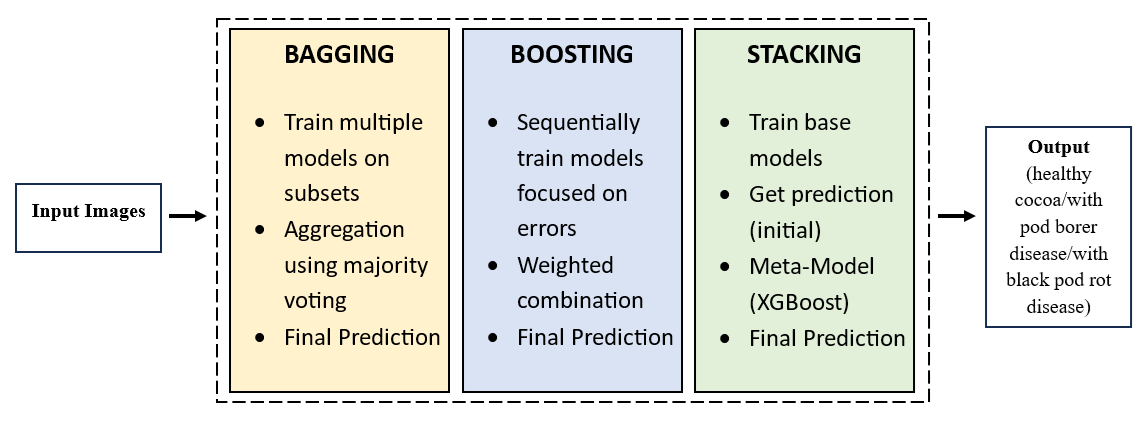}
	\caption{The conceptual framework of the proposed solution.}
	\label{fig:una}
\end{figure*}

Recent developments in computer vision and machine learning have shown potential in automating plant disease diagnosis. In particular, \textbf{transfer learning} and \textbf{ensemble methods} have demonstrated improved performance in classification tasks. Ensemble learning combines the predictions of multiple base classifiers to reduce variance, enhance stability, and improve accuracy \cite{dietterich2000ensemble, brown2010good, rokach2010pattern}. However, its effectiveness relies on the diversity and quality of the base models used.

Despite its promise, ensemble learning has not been widely explored for cocoa pod disease classification. Prior studies often utilized single classifiers or traditional machine learning techniques, which may fall short in addressing the complexity of disease patterns and visual variability \cite{zhang2018learning,thakur2022trends}.

This study proposes an ensemble-based framework that integrates \textbf{transfer learning models}—such as VGG, ResNet, and others—as weak learners within \textbf{bagging}, \textbf{boosting}, and \textbf{stacking} schemes. By fine-tuning these pre-trained CNNs and systematically combining their predictions, the research aims to identify the optimal ensemble configuration for maximizing classification accuracy and reliability in cocoa pod disease detection.

To address the limitations of single-model classifiers and advance automated disease detection in cacao cultivation, this research leverages ensemble learning techniques in conjunction with weak learners derived from fine-tuned transfer learning models. The framework aims to improve the accuracy and robustness of cocoa pod disease classification by evaluating the performance of different ensemble strategies and determining the most effective method for maximizing predictive performance.

\section{Related work}\label{sec2}
Ensemble learning has gained prominence in improving classification accuracy by combining multiple base models to reduce variance and bias \cite{dietterich2000ensemble, rokach2010pattern}. The three main types—bagging, boosting, and stacking—operate either in parallel or sequentially. Bagging methods, such as Random Forest and Extra Trees, aggregate predictions from models trained on bootstrap samples, while boosting methods like AdaBoost, Gradient Boosting, and XGBoost sequentially refine model errors. Stacking leverages meta-learners to combine outputs from diverse base models \cite{lersteau2021solving, mienye2022survey}.

In agriculture, ensemble methods have been successfully applied to plant disease classification. \cite{ennouni2021analysis} proposed an ensemble classifier combining SqueezeNet, VGG, InceptionV3, and DeepLoc for fruit disease detection, achieving up to 99\% accuracy. \cite{govardhan2019diagnosis} classified tomato leaf diseases using Random Forest, SVM, and MLP, attaining over 92\% accuracy.

Similarly, stacking-based ensemble classifiers have proven effective in diagnosing fruit tree diseases, with \cite{li2021fruit} reporting a test accuracy of 97.34\% on a 10,000-image dataset. \cite{kayaalp2024deep} achieved 100\% accuracy in cherry classification using maximum voting over deep models. \cite{nader2022grape} applied ensemble deep learning (VGG16, VGG19, Xception) to grape leaf disease classification, achieving 99.82\% accuracy.

\cite{fenu2023classification} utilized ensemble CNNs (EfficientNet, InceptionV3, MobileNetV2, VGG19) with bagging and weighted averaging for pear disease detection, demonstrating superior performance under real-field conditions. \cite{ennouni2021analysis} employed hard and soft voting strategies using five deep models for plant disease classification, with hard voting nearing 100\% accuracy. \cite{dewangan2022leaf} implemented an optimized ensemble to detect wheat diseases, achieving classification accuracy up to 98.25\%.

While ensemble learning is well-explored across various crops, its application to cocoa pod disease remains limited. This study builds on the demonstrated success of ensemble frameworks in agriculture by combining transfer learning with ensemble strategies to improve the accuracy and reliability of cocoa pod disease classification.

\begin{figure}[t]
	\centering
	\includegraphics[width=\columnwidth,height=6cm,keepaspectratio]{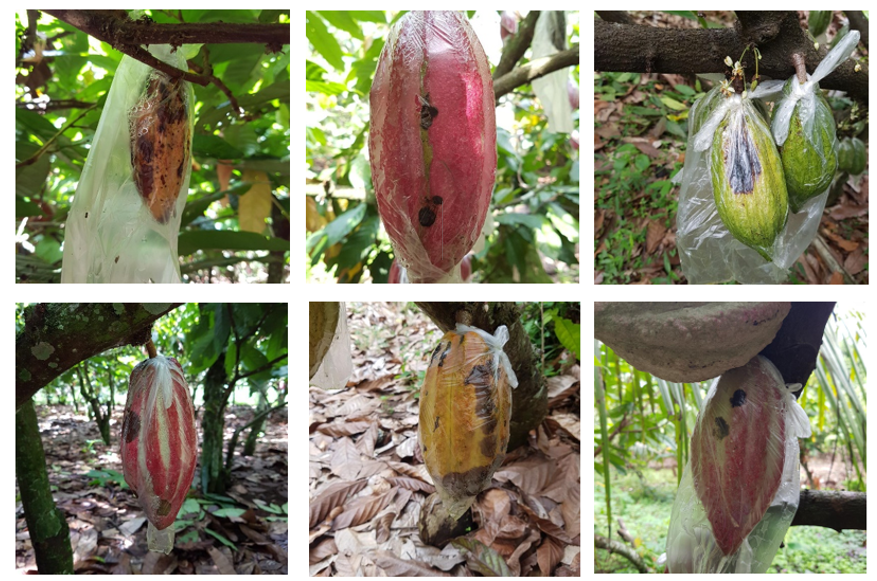}
	\caption{Sample cocoa pod images affected by Pod Borer disease.}
	\label{fig:2}
\end{figure}

\section{Materials and methods}\label{sec4}
The focus of this study is to determine the most effective ensemble learning method for cocoa pod disease classification by combining fine-tuned transfer learning models. Considering variations in pod appearance, lighting conditions, and disease severity, the study evaluates ensemble techniques such as bagging, boosting, and stacking. Classification performance is measured using standard quantitative metrics to assess accuracy, robustness, and practical effectiveness in real-world agricultural settings.

\subsection{Dataset}
The dataset used in this study, titled \textit{Cacao Disease}, was manually obtained from the Kaggle platform. It comprises approximately 4,300 images of cocoa pods with a resolution of $1080 \times 1080$ pixels, categorized into three classes: \textit{Black Pod Rot}, \textit{Healthy}, and \textit{Pod Borer}. 

To ensure consistency and improve model performance, the images were pre-processed by resizing them to a standard resolution of $240 \times 240$ pixels. In addition, data augmentation techniques such as random rotation and horizontal flipping were applied to increase dataset diversity and enhance the model’s ability to generalize across different orientations and visual perspectives of cocoa pods.

\begin{figure}[t]
	\centering
	\includegraphics[width=\columnwidth,height=6cm,keepaspectratio]{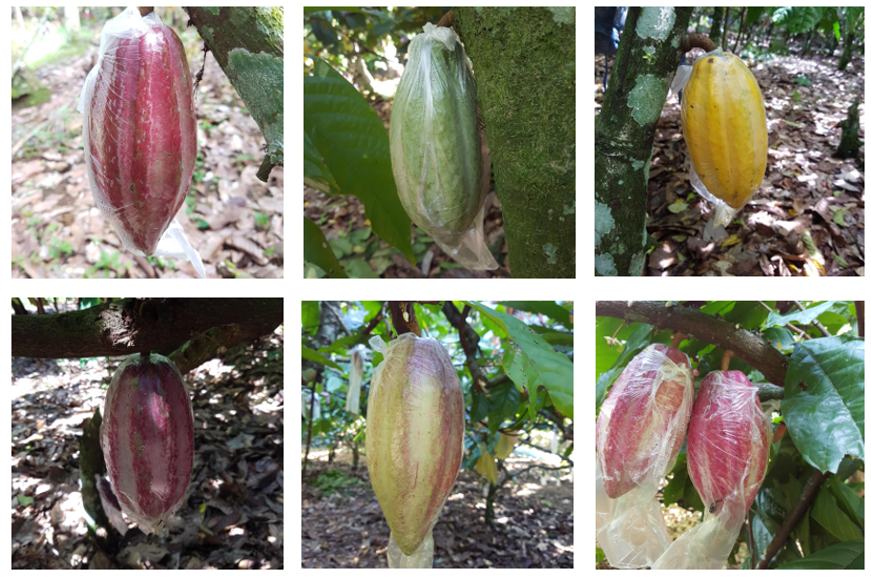}
	\caption{Sample of healthy cocoa pod images.}
	\label{fig:3}
\end{figure}

\begin{figure}[t]
	\centering
	\includegraphics[width=\columnwidth,height=6cm,keepaspectratio]{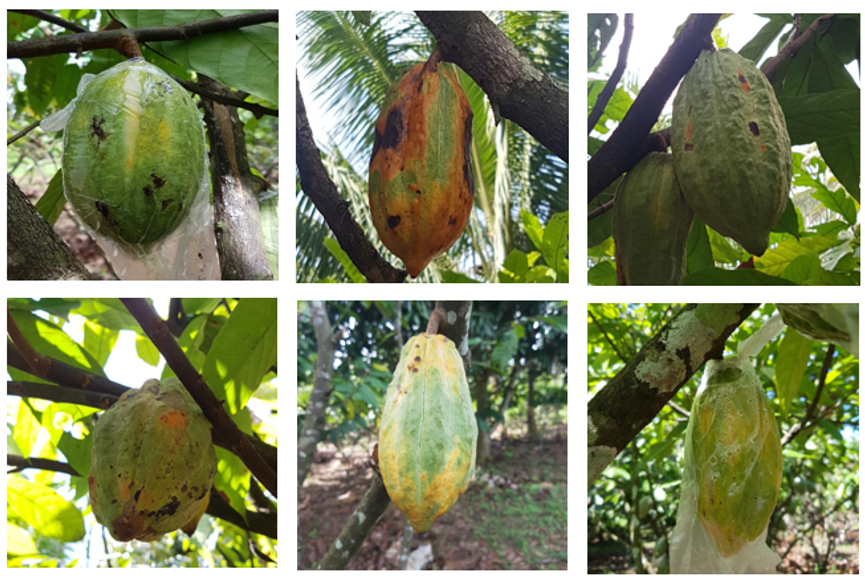}
	\caption{Sample cocoa pod images affected by Black Pod Rot disease.}
	\label{fig:4}
\end{figure}

Following preprocessing and augmentation, the dataset was balanced to contain 2,000 images per class, all stored in \texttt{.jpg} format. Representative samples from each class are shown in Figures~\ref{fig:2},~\ref{fig:3}, and~\ref{fig:4}.

\subsection{Base Models}
This study employs established convolutional neural network (CNN) architectures—VGG16 \cite{simonyan2014very}, VGG19 \cite{simonyan2014very}, InceptionV3 \cite{szegedy2016rethinking}, Xception \cite{chollet2017xception}, ResNet50 \cite{he2016deep}, and ResNet101~\cite{he2016deep}—as base learners within the ensemble framework. These models were selected due to their proven performance in image classification tasks and their capacity to learn complex features and visual patterns. The use of these pre-trained networks aims to enhance the accuracy and robustness of the cocoa pod disease classification system by leveraging their transfer learning capabilities.

\subsection{Ensemble Learning Techniques}
In this study, ensemble learning techniques—bagging, boosting, and stacking—are implemented to improve the performance of transfer learning models for cocoa pod disease classification. These methods aim to enhance prediction accuracy and robustness by leveraging multiple base learners and aggregating their outputs through different ensemble strategies.

\subsubsection{Bagging}

Bagging (Bootstrap Aggregating) is a parallel ensemble technique where multiple base learners are trained independently on different subsets of the dataset generated through bootstrapping. This reduces variance and mitigates overfitting. For each new sample, predictions from all learners are aggregated using majority voting.

Given a dataset $D = \{ (x_1, y_1), (x_2, y_2), \dots, (x_n, y_n) \}$, $m$ bootstrap samples $D_i$ are generated by random sampling with replacement. Each base model $M_i$ is trained on a corresponding $D_i$. For a test sample $x$, the final predicted class $\hat{y}$ is obtained via:

\begin{equation}
	\hat{y} = \arg\max_{c_j \in C} \sum_{i=1}^{m} \mathbb{I}[M_i(x) = c_j]
	\label{eq:bagging_vote}
\end{equation}

where $\mathbb{I}[\cdot]$ is the indicator function and $C$ is the set of possible classes.

\subsubsection{Boosting}

Boosting is a sequential ensemble method where each base learner is trained to correct the errors of its predecessor. The models are trained iteratively, and their predictions are combined using a weighted majority scheme. Boosting focuses on difficult samples by assigning them higher weights in subsequent rounds.

The boosting process follows:
\begin{enumerate}
	\item Initialize weights $w_i = \frac{1}{n}$ for all training instances.
	\item For $t = 1$ to $T$ (number of rounds):
	\begin{enumerate}
		\item Train base learner $M_t$ on weighted data.
		\item Compute weighted error $\varepsilon_t$:
		\begin{equation}
			\varepsilon_t = \sum_{i=1}^{n} w_i \cdot \mathbb{I}[M_t(x_i) \neq y_i]
			\label{eq:boosting_error}
		\end{equation}
		\item Compute model weight:
		\begin{equation}
			\alpha_t = \frac{1}{2} \ln\left(\frac{1 - \varepsilon_t}{\varepsilon_t}\right)
			\label{eq:boosting_weight}
		\end{equation}
		\item Update sample weights:
		\begin{equation}
			w_i \leftarrow w_i \cdot \exp\left(-\alpha_t y_i M_t(x_i)\right)
			\label{eq:boosting_update}
		\end{equation}
		\item Normalize weights so that $\sum w_i = 1$.
	\end{enumerate}
\end{enumerate}

The final prediction is computed as:
\begin{equation}
	\hat{y} = \text{sign} \left( \sum_{t=1}^{T} \alpha_t M_t(x) \right)
	\label{eq:boosting_final}
\end{equation}

\subsubsection{Stacking}

Stacking is a layered ensemble technique that combines predictions from multiple base models using a meta-model. Base learners are trained on the original training data, and their outputs are used as input features for a higher-level model that learns how to best combine them.

Let $M_1, M_2, \dots, M_k$ be base models. For each training instance $x_i$, meta-features are generated as $Z_i = [M_1(x_i), M_2(x_i), \dots, M_k(x_i)]$. A meta-model $M_{\text{meta}}$ is then trained using the feature matrix $Z = \{Z_1, Z_2, \dots, Z_n\}$ and original labels $Y = \{y_1, y_2, \dots, y_n\}$. The final prediction for a new sample $x$ is:

\begin{equation}
	\hat{y} = M_{\text{meta}}([M_1(x), M_2(x), \dots, M_k(x)])
	\label{eq:stacking_final}
\end{equation}

Stacking enables the system to leverage the strengths of different base models and reduce generalization error.

\subsection{Evaluation Metrics}

After fine-tuning and validating the pre-trained models, their performance is assessed on the test dataset using standard classification metrics: accuracy, precision, recall, and F1-score. These metrics provide a comprehensive evaluation of the model's predictive capabilities across all cocoa pod disease classes.

\begin{itemize}
	\item \textbf{Accuracy} measures the proportion of correctly predicted instances over the total number of predictions. It provides an overall indication of model performance:
	
	\begin{equation}
		\text{Accuracy} = \frac{TP + TN}{TP + TN + FP + FN}
		\label{eq:accuracy}
	\end{equation}
	
	where:
	\begin{itemize}
		\item $TP$ = True Positives
		\item $TN$ = True Negatives
		\item $FP$ = False Positives
		\item $FN$ = False Negatives
	\end{itemize}
	
	\item \textbf{Precision} indicates the correctness of positive predictions, i.e., the proportion of predicted positive cases that are truly positive:
	
	\begin{equation}
		\text{Precision} = \frac{TP}{TP + FP}
		\label{eq:precision}
	\end{equation}
	
	\item \textbf{Recall} (also known as sensitivity) measures the model’s ability to correctly identify actual positive cases:
	
	\begin{equation}
		\text{Recall} = \frac{TP}{TP + FN}
		\label{eq:recall}
	\end{equation}
	
	\item \textbf{F1-Score} provides a harmonic mean of precision and recall, balancing both metrics into a single value:
	
	\begin{equation}
		\text{F1-Score} = \frac{2 \cdot \text{Precision} \cdot \text{Recall}}{\text{Precision} + \text{Recall}}
		\label{eq:f1score}
	\end{equation}
\end{itemize}

These metrics are especially relevant in multi-class classification tasks such as cocoa pod disease detection, where both over- and under-predictions of disease types can significantly affect model reliability.

\section{Experiments}

\subsection{Cocoa pod disease classification results}
\begin{figure}[h]
	\centering
	\includegraphics[width=\columnwidth,height=6cm,keepaspectratio]{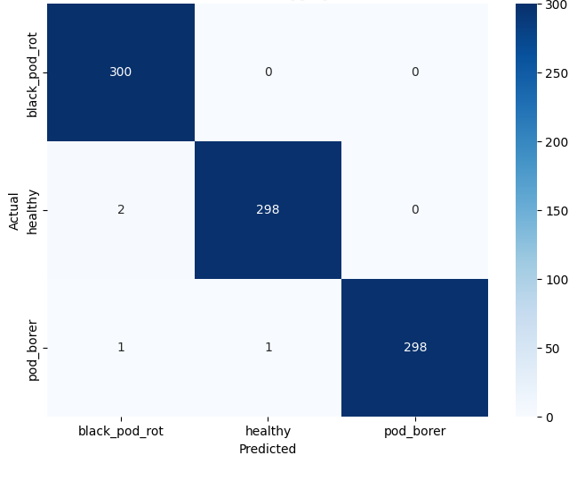}
	\caption{Confusion Matrix - Bagging (Majority Voting) Ensemble.}
	\label{fig:6}
\end{figure}

Table 1 and the corresponding confusion matrix in Figure \ref{fig:6}, demonstrate the high performance of the bagging ensemble with majority voting on the cocoa pod disease classification task. The model achieved an overall accuracy of 100\%, with only four misclassifications out of 900 test samples: two healthy misclassified as black pod rot, and one pod borer each misclassified as black pod rot and healthy. Despite these minor errors, the precision, recall, and F1-scores remained at or near 1.00 across all classes, indicating excellent robustness and class balance.
\begin{center}
	\captionof{table}{Classification Report - Bagging (Majority Voting) Ensemble}
	\includegraphics[width=\columnwidth,height=19cm,keepaspectratio]{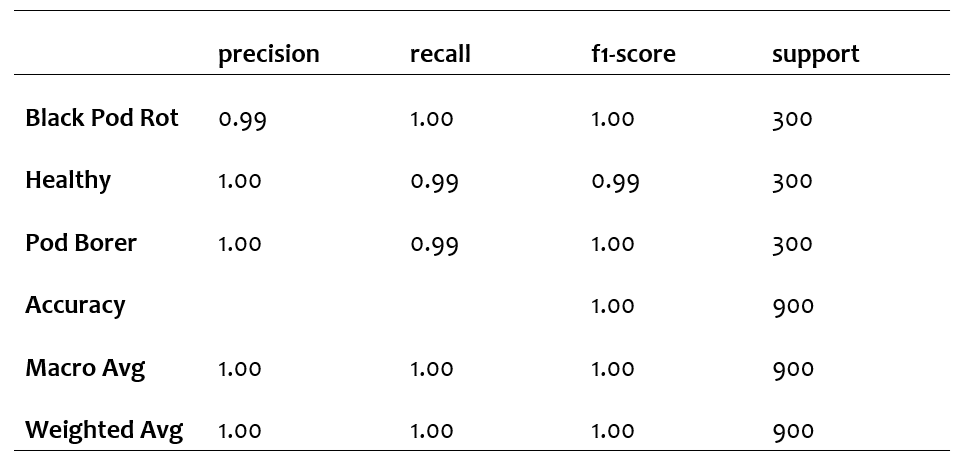}
\end{center}

\begin{figure}[h]
	\centering
	\includegraphics[width=\columnwidth,height=6cm,keepaspectratio]{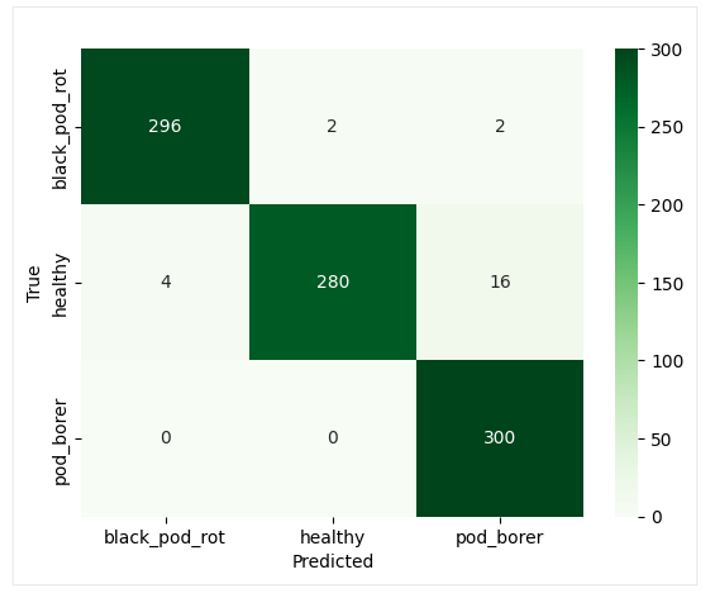}
	\caption{Confusion Matrix - Boosting Ensemble.}
	\label{fig:8}
\end{figure}

Table 2 and the associated confusion matrix in Figure \ref{fig:8}, summarize the performance of the Boosting ensemble method. The model achieved an overall accuracy of 97\%, with class-level F1-scores of 0.99 for Black Pod Rot, 0.96 for Healthy, and 0.97 for Pod Borer. The confusion matrix reveals minor misclassifications, including 4 Healthy samples incorrectly predicted as Black Pod Rot and 16 as Pod Borer. While Pod Borer was classified perfectly, Healthy class exhibited the most confusion, slightly reducing recall. Nonetheless, macro and weighted averages of 0.97 across all metrics demonstrate the model’s strong and balanced performance across classes.
\begin{center}
	\captionof{table}{Classification Report - Boosting Ensemble}
	\includegraphics[width=\columnwidth,height=19cm,keepaspectratio]{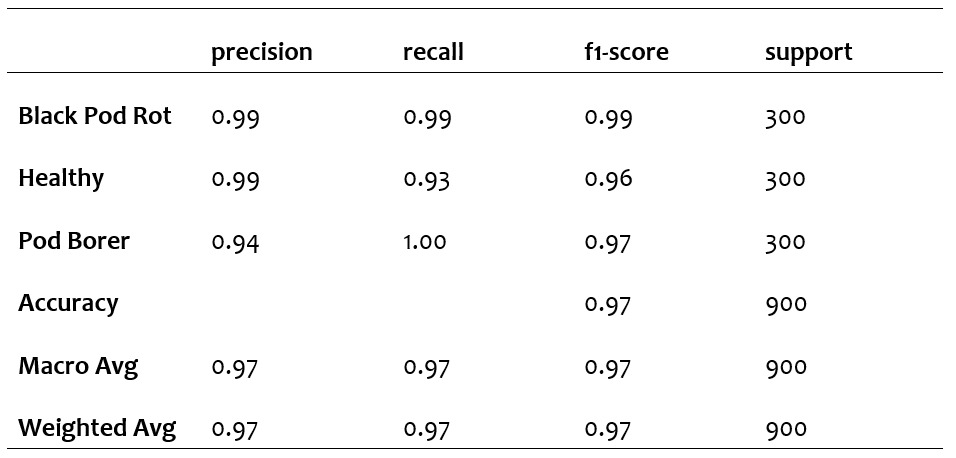}
\end{center}

\begin{figure}[h]
	\centering
	\includegraphics[width=\columnwidth,height=6cm,keepaspectratio]{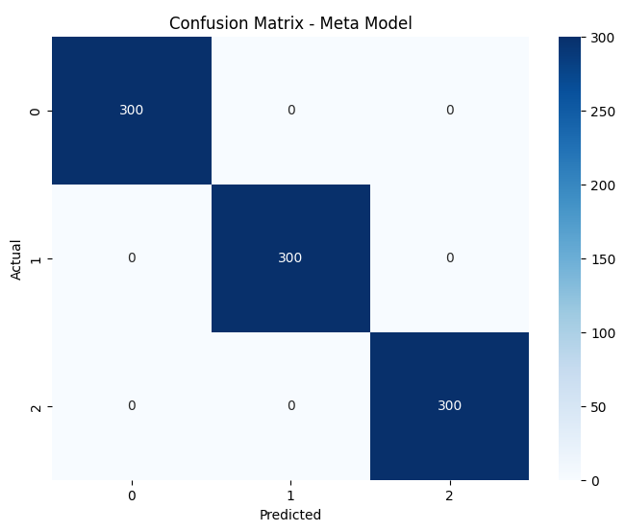}
	\caption{Confusion Matrix - Stacking Ensemble.}
	\label{fig:10}
\end{figure}

Table 3 and the corresponding confusion matrix in Figure \ref{fig:10}, present the evaluation results of the stacking ensemble model. Despite achieving a perfect confusion matrix with all 900 test samples correctly classified, the class-wise F1-scores reveal some inconsistencies. While Black Pod Rot was predicted with high consistency (F1 = 0.93), Healthy had a lower precision (0.83), and Pod Borer showed a lower recall (0.85). These discrepancies suggest internal class imbalance or meta-feature misalignment despite perfect final predictions. The model achieved a macro and weighted average of 0.92 across all metrics, indicating good but not superior generalization compared to bagging or boosting.
\begin{center}
	\captionof{table}{Classification Report - Stacking Ensemble}
	\includegraphics[width=\columnwidth,height=19cm,keepaspectratio]{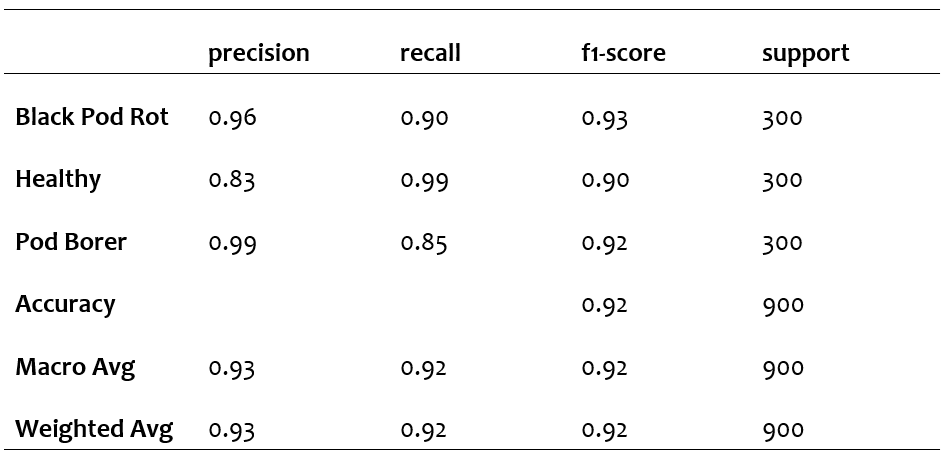}
\end{center}

\subsection{Model performance across ensemble methods}
\begin{figure*}[h]
	\centering
	\includegraphics[width=16cm,height=20cm,keepaspectratio]{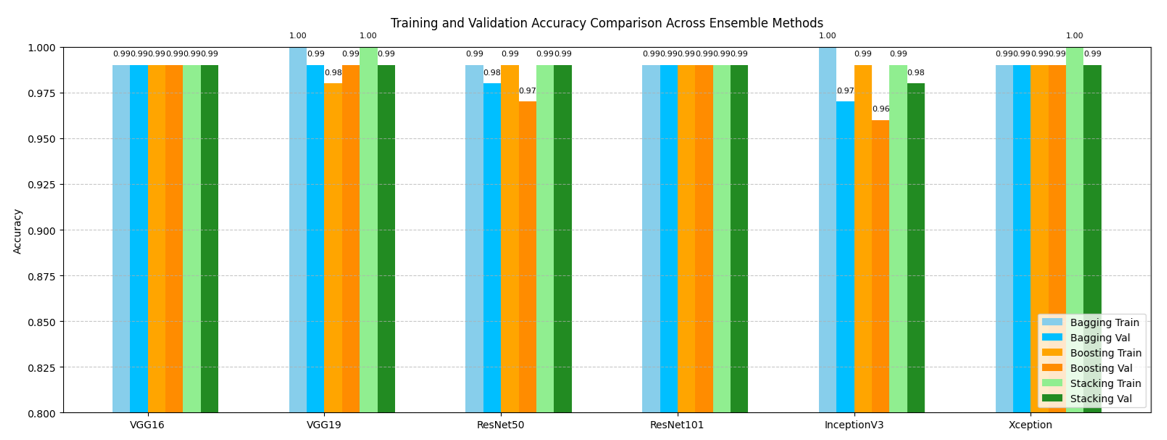}
	\caption{Training and validation accuracy of pre-trained CNN models (VGG16, VGG19, ResNet50, ResNet101, InceptionV3, and Xception) using Bagging, Boosting, and Stacking ensemble methods.}
	\label{fig:11}
\end{figure*}

Figure~\ref{fig:11} illustrates the training and validation accuracy of six pre-trained CNN models (VGG16, VGG19, ResNet50, ResNet101, InceptionV3, and Xception) under three ensemble learning strategies: \textit{Bagging}, \textit{Boosting}, and \textit{Stacking}. Across all architectures, Bagging consistently achieved near-perfect validation accuracy ($\geq 0.99$), indicating strong generalization and model stability.

Boosting also performed well, although minor drops were observed in VGG19 and ResNet50, where validation accuracy dipped to 0.98 and 0.97, respectively—suggesting slight overfitting or sensitivity to sample weighting. Stacking maintained competitive performance with validation accuracies closely matching Bagging, particularly for deeper architectures like ResNet101 and Xception.

Overall, Bagging delivered the most consistent performance across all models, with minimal variance between training and validation scores. Boosting showed excellent training accuracy but slight drops in generalization. Stacking provided robust results, though with more variability depending on the base architecture. These trends highlight the superior stability of Bagging for cocoa pod disease classification tasks, especially when paired with deep pre-trained models.

\begin{figure*}[h]
	\centering
	\includegraphics[width=13cm,height=20cm,keepaspectratio]{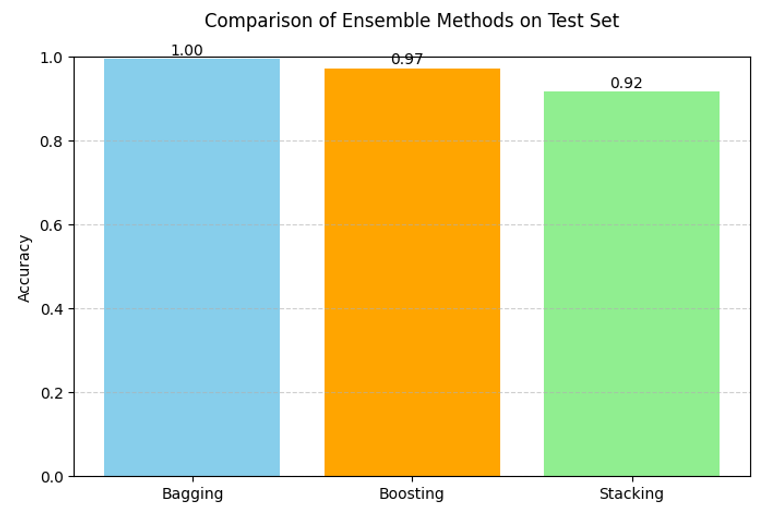}
	\caption{Comparison of test set accuracy for Bagging, Boosting, and Stacking ensemble methods applied to cocoa pod disease classification.}
	\label{fig:12}
\end{figure*}

Figure~\ref{fig:12} compares the overall test set accuracy of the three ensemble learning methods applied to cocoa pod disease classification. Bagging achieved a perfect test accuracy of 1.00, clearly outperforming both Boosting (0.97) and Stacking (0.92). These results confirm Bagging’s superior generalization ability, as previously observed in the validation phase (see Figure~\ref{fig:11}), where it consistently yielded high and stable performance across all backbone models.

Boosting maintained strong performance but exhibited slight degradation, likely due to its sensitivity to noisy samples or class imbalance, which was reflected in both validation and test stages. Stacking, despite showing reasonable validation scores, had the lowest test accuracy, suggesting it may have overfit to the meta-features or lacked sufficient diversity among base learners.

Together, these findings underscore Bagging as the most robust and reliable ensemble method for cocoa pod disease classification in this study.

\section{Conclusion}
This study investigated the effectiveness of combining transfer learning and ensemble learning techniques for cocoa pod disease classification. By fine-tuning pre-trained CNN models—VGG16, VGG19, ResNet50, ResNet101, InceptionV3, and Xception—and integrating them into Bagging, Boosting, and Stacking frameworks, we evaluated their performance on a balanced and augmented dataset containing three disease classes. Among the ensemble methods, Bagging demonstrated the highest classification accuracy (100\%), followed by Boosting (97\%) and Stacking (92\%). The results highlight Bagging’s superior generalization capabilities and consistent performance across different architectures. Boosting offered competitive results but showed sensitivity to class imbalance, while Stacking introduced variability likely due to the meta-model's reliance on base learner outputs.

Overall, the proposed ensemble-based approach—particularly when leveraging Bagging—proves to be an effective and robust solution for automated cocoa pod disease detection. Future work may explore lightweight deployment for real-time UAV-based monitoring and integration of additional visual features to improve disease differentiation under real-field conditions.

\bmhead{Acknowledgements}
The authors would like to extend their gratitude to the CCIS Network Team, Caraga State University, for providing access to their high-performance computing resources, which were essential for the development and testing of the software. Appreciation is also extended to all support systems and other relevant institutional resources that facilitated the successful completion of this research.

\section*{Declarations}

\paragraph*{Conflicts of interest/Competing interests}
\ No conflict of interest being declared.

\bibliography{sn-bibliography}

\end{document}